\newif\ifprl
\prlfalse

\documentclass[prl,superscriptaddress,amsmath,amssymb,twocolumn]{revtex4-1}
\pdfoutput=1

\usepackage{amsthm}
\usepackage{graphicx}
\usepackage[usenames,dvipsnames]{color}

\usepackage{environ}  

\ifprl 
\newcommand{\mysection}[1]{{\bf #1} ---}
\NewEnviron{equ}{%
\(
  \BODY
\)
}
\else
\newcommand{\mysection}[1]{\section{#1}}
\NewEnviron{equ}{%
\begin{equation}
  \BODY
\end{equation}
}
\fi

\newcommand{\mc}[1]{\mathcal #1}

\newcommand{\ave}[1]{\langle #1 \rangle}

\newcommand{\tr}{{\rm Tr}\,}

\newcommand{\be}{\begin{equation}}
\newcommand{\ee}{\end{equation}}
\newcommand{\bs}{\begin{split}}
\newcommand{\es}{\end{split}}

\theoremstyle{plain}

\theoremstyle{definition}

\begin{document}

\title{Inferring relevant features: from QFT to PCA}
\author{C\'edric B\'eny}
\address{Department of Applied Mathematics, Hanyang University (ERICA), 55 Hanyangdaehak-ro, Ansan, Gyeonggi-do, 426-791, Korea.}
\address{Institut f\"ur Theoretische Physik, Leibniz Universit\"at Hannover, Appelstra{\ss}e 2, 30167 Hannover, Germany}

\begin{abstract}
In many-body physics, renormalization techniques are used to extract aspects of a statistical or quantum state that are relevant at large scale, or for low energy experiments. Recent works have proposed that these features can be formally identified as those perturbations of the states whose distinguishability most resist coarse-graining. 
Here, we examine whether this same strategy can be used to identify important features of an unlabeled dataset. 
This approach indeed results in a technique very similar to kernel PCA (principal component analysis), but with a kernel function that is automatically adapted to the data, or ``learned''. We test this approach on handwritten digits, and find that the most relevant features are significantly better for classification than those obtained from a simple gaussian kernel.
\end{abstract}

\maketitle

Brains evolved to model reality, so as to make predictions useful for survival. Because sciences such as physics expand and formalize this activity, their methods may contain clues on how to automatize natural intelligence. Renormalization may be a particularly pertinent technique to examine for that purpose, because it provides an explicit methodology for the creation of effective models of complex phenomena~\cite{wilson1983}.

For instance, connections between hierarchical neural networks and renormalization were established in Refs~\cite{beny2013b,mehta2014,gan2017}. Here, we establish a different type of connection between the two. 

Renormalization can be thought of as follows: given some general assumptions about a quantum state (e.g., it is close to the ground states of a free field theory), and some specification of what we cannot confidently observe (small scale or high energy fluctuations), then what parameters of the state are most relevant to us experimentally?

Recent works~\cite{Machta2013,beny2013} recognized that this strategy resemble one of dimensional reduction in data analysis, and proposed (quantum) Fisher information metrics, which are measures of statistical distinguishability, as the yardsticks with which relevance of state parameters is to be measured.

Here, we apply the specific framework proposed in Refs.~\cite{beny2013,beny2015a} to machine-learning, specifically unsupervised feature extraction. This results in a technique similar to kernel PCA (principal component analysis)~\cite{scholkopf1998}, where the ``kernel'' (Eq.~\eqref{kernel}) is adapted to the data. 

This technique also has a natural Bayesian interpretation as follows: it finds the parameters which can be most faithfully inferred from the coarse-grained data.

We test our method on a subset of the MNIST handwritten digits dataset, and find that it automatically extracts most of the relevant information about the identity of the digits already in the first few most significant components.  

A connection between renormalization and PCA was explored also in Ref.~\cite{bradde2017}. In that work, the authors study the relation between renormalization and the idea of cutting off small eigenvalue components of the covariance matrix between field operators. There is indeed a connection between this idea and our approach due to the fact that the covariance matrix is essentially also the Fisher information metric in that context~\cite{beny2015}. However, a central difference is the fact the we consider the relative {\em degradation} of distinguishability due to coarse-graining.

\section{Framework}

The framework requires a rough specification of those aspects of the system that we cannot observe, or do not care about. We model this by a channel $\mc N$ (a trace-preserving completely-positive linear map). In the context of renormalization, this channel would implement a coarse-graining which purposefully degrades information about the small scale details of a quantum field (without however erasing them entirely). 
The other required input is a specific density matrix $\rho$ which plays a role similar to a null hypothesis or a Bayesian prior (more on this below). 

Although we use the quantum formalism for now, this applies as is for classical probability theory: one needs just assume that $\rho$ is diagonal and that $\mc N$ maps diagonal states to diagonal states---it is then a stochastic map. 

Let's consider any parameterization $\rho_s$ such that $\rho_0 = \rho$, where $s \in \mathbb R^N$ for concreteness. We measure the {\em relevance} of a parameter vector $s$ by how distinguishable $\mc N(\rho_{s})$ is from $\mc N(\rho_0)$, compared to the distinguishability between $\rho_s$ and $\rho_0$, to first order in $s$.

We remark that this concept of relevance can be thought of as a refinement of that used in quantum field theory. For instance, consider a QFT Hamiltonian $H$ and a translation-invariant interaction term $V$. If $\rho_s$ denotes the ground state of $H + sV$, and the channel $\mc N$ erases details up to a spatial scale $\sigma$, then one may discuss whether the coarse-grained distinguishability {\em density} of a change in $s$ depends on $\sigma$, if the density is evaluate with respect to the scale $\sigma$ as well. The coarse-grained distinguishability density of the parameter $s$ may then increase or decrease as a function of $\sigma$. If it increases then $V$ is relevant in the usual sense. This is explained in detail in Section VII A of Ref.~\cite{Beny2015b}.




Since our approach is independent of any choice of parameterization, then instead of measuring the relevance of a parameter vector $s$, we can directly consider the relevance of a tangent vector represented as the traceless self-adjoint operator $X = \partial_s \rho_s |_{s=0}$. We then have $\rho_s = \rho + s X + \mc O(s^2)$. We will think of these tangent vectors as {\em features}.

A measure of distinguishability $D(\cdot,\cdot)$ that is adequately differentiable induces a bilinear form on features (tangent vectors at $\rho$) as
\begin{equ}
D(\rho + sX, \rho) = s^2 \ave{X,X}_\rho + \mc O(s^4).
\end{equ}
For classical probability theory, the only consistent possibility is the Fisher information metric $\ave{X,X}_\rho = \tr(X^2 \rho^{-1})$.

The relevance of a feature $X$ is then given by
\begin{equ}
\eta(X) = \frac{\ave{\mc N(X),\mc N(X)}_{\mc N(\rho)}}{\ave{X,X}_\rho},
\end{equ}
which is between $0$ and $1$.

The more relevant features are the eigenvectors of larger eigenvalues for the superoperator $\mc N^*_\rho \mc N$ (the composition of $\mc N$ and $\mc N^*_\rho$), where $\mc N^*_\rho$ is the adjoint of $\mc N$ with respect to the bilinear form $\ave{\cdot,\cdot}_\rho$ (or rather the transpose since the tangent space is real), i.e., it is defined by
\begin{equ}
\ave{X, \mc N^*_\rho(Y)}_{\rho} = \ave{\mc N(X), Y}_{\mc N(\rho)},
\end{equ}
so that, indeed,
\begin{equ}
\eta(X) = \frac{\ave{X,\mc N_\rho^* \mc N(X)}_{\rho}}{\ave{X,X}_\rho}.
\end{equ}
In the classical setting, where everything commutes, we find
\begin{equation}
\label{scadj}
\mc N_\rho^*(Y) = \rho\, \mc N^\dagger(\mc N(\rho)^{-1} Y),
\end{equation}
where $\mc N^\dagger$ is the Hilbert-Schmidt adjoint of $\mc N$: $\tr(\mc N(\rho)A) = \tr(\rho \mc N^\dagger(A))$ for all states $\rho$ and operators $A$.

This approach is very similar that that which is proposed in Ref.~\cite{Machta2013} when restricted to a classical statistical setting. However, the quantity that these authors diagonalize, interpreted within our setting, is not quite equivalent to the linear map $\mc N_\rho^* \mc N$, but rather the matrix
\begin{equ}
M_{ij} =\ave{X_i,\mc N_\rho^* \mc N(X_j)}
\end{equ}
for some specific choice of the variables $X_1, \dots, X_n$. But this is of course equivalent if the variables $X_i$ form an orthonormal basis in terms of the Fisher information metric, which will turn out to be the case in our application below.

\section{Relevant feature extraction}

Suppose we have a classical data set of points $x_1, \dots, x_n$ from some configuration space, say $\mathbb R^m$. Let us imagine that they are samples from an unknown probability distribution over that space. As the state $\rho$, we choose a very rough estimate of this probability distribution, namely the ``empirical distribution''
\begin{equ}
x \; \longmapsto \; \rho_x = \frac 1 n \sum_{i=1}^n \delta_{x_i,x},
\end{equ}
where $x \in \mathbb R^m$ and $\delta$ is the Kronecker delta. (We now use the classical notation where $\rho$ is just a positive real function).

In order to apply our scheme, we also need a stochastic map $\mc N$ which defines those aspects of the configuration space that we deem unimportant. For now, let us consider a generic case, mapping the probability distribution $\rho$ to $\mc N(\rho)$ given by
\begin{equ}
\mc N(\rho)_y = \int p(y|x) \rho_x,
\end{equ}
where $p(y|x)$ are the conditional (transition) probabilities. 

With this notation, we can now see that the transpose map $\mc N_\rho^*$ represents Bayesian inference on the conditional probability $p(y|x)$ with prior $\rho$. Indeed, using Eq.~\eqref{scadj} with the classical notation, we obtain
\begin{equ}
\mc N^*_\rho (\delta_z)_x = \frac{p(z|x) \rho(x)}{\int dx \,p(z|x) \rho(x)}.
\end{equ}

Recall the we want to compute eigenvectors of the superoperator $\mc N^*_\rho \mc N$. 
In order to express it in a way that can be diagonalized, we employ a standard trick (as in kernel PCA) of considering only the subspace of probability distribution spanned by the empirical pure states $\delta_{x_i}$, $i=1, \dots, n$. The dimension $n$ of this subspace is just the number of sample points.  

Since the features $\delta_{x_i}$ or orthogonal in the Fisher metric:
\begin{equ}
\ave{\delta_{x_i},\delta_{x_j}}_\rho = n \, \delta_{ij},
\end{equ}
the components of the linear map $\mc N^*_\rho \mc N$ in the span of the samples are 
\begin{equ}
\label{kernel}
\begin{split}
K_{ij} &= \frac 1 n \ave{\delta_{x_i},\mc N_\rho^*\mc N(\delta_{x_j})}_{\rho} =\frac 1 n \ave{\mc N(\delta_{x_i}),\mc N(\delta_{x_j})}_{\mc N(\rho)} \\
&= \int_{\mathbb R^m} dx\, \frac{p(x|x_i) p(x|x_j)}{\sum_{k=1}^n p(x|x_k)}.
\end{split}
\end{equ}

From the last expression in this formula, we immediately see an alternative interpretation: $K_{ij}$ is the average probability of inferring that the input of the channel was $x_i$ when it actually was $x_j$ (using Bayesian inference with prior $\rho$). 

In principle, the $k$ largest eigenvectors of the $n$-by-$n$ matrix $K$ can then be determined numerically. A given eigenvector with components $v^1,\dots,v^n$ corresponds to the concrete feature (tangent vector)
\begin{equ}
X = \frac 1 {\sqrt n} \sum_{i=1}^n v^i \delta_{x_i}.
\end{equ}

The $k$ most relevant features $X_1, \dots, X_k$ can be used for instance to compress a new data point $z$ by keeping only its components with respect to these vectors. Note that, since $\mc N^*_\rho \mc N$ is self-adjoint in the Fisher metric, its eigenvectors $X_i$ are automatically orthogonal in the metric.
If we also normalize them, then the $k$ most relevant components of a new data point $z$ are given by
\begin{equ}
\tilde z^j := \frac 1 {\sqrt n} \ave{\delta_z, X_j}_{\rho}. 
\end{equ}
We cannot however use the expansion of $X_j$ in terms of the vectors $\delta_{x_i}$, because they have zero overlap with $\delta_z$ (this is possible because these vectors do not form a complete family). However, we can rewrite this expression as
\begin{equ}
\label{proj0}
\begin{split}
\tilde z^j &= \frac 1 {\sqrt n \, \eta_j} \ave{\delta_z, \mc N^*_\rho \mc N(X_j)}_{\rho}\\
&= \frac 1 {n \,\eta_j} \sum_i X_j^i \ave{\delta_z, \mc N^*_\rho \mc N(\delta_{x_i})}_{\rho}\\
&= \frac 1 {\eta_j} \sum_i X_j^i \int dx\, \frac{p(x|z)p(x|x_i)}{\sum_l p(x|x_l)}
\end{split}
\end{equ}
where $\eta_j$ is the eigenvalue for the eigenvector $X_j = \frac 1 {\sqrt n} \sum_i X_j^i \delta_{x_i}$, i.e., $\mc N^*_\rho \mc N(X_j) = \eta_j X_j$.

An alternative is to assume instead that the point $z$ was sampled from the image of the true state under the channel $\mc N$. Then we can project it on the images $\mc N(X_j)$ instead of on $X_j$ directly:
\begin{equ}
\label{proj}
\begin{split}
z^j &:= 
 \frac 1 {\sqrt n} \ave{\delta_z, \mc N(X_j)}_{\mc N(\rho)}\\ 
&= \frac 1 n \sum_i X_j^i \ave{\delta_z, \mc N(\delta_{x_i})}_{\mc N(\rho)}\\
&= \sum_i X_j^i \, \frac{p(z|x_i)}{\sum_l p(z|x_l)}.\\
\end{split}
\end{equ} 
This is much faster to evaluate as it does not require any integration. Moreover, it performs much better as shown in Fig.~\ref{classification}.

\section{Numerical approximation}

In order to evaluate the integral defining $K_{ij}$ in Eq.~\eqref{kernel}, we observe that it can be thought of as performing an average of the function
\begin{equ}
f_j(x) = \frac{p(x|x_j)}{\sum_k p(x|x_k)}
\end{equ} 
over the probability distribution $p(x|x_i)$. 
This average can be evaluated for instance by Montecarlo sampling, depending on how the conditional probabilities $p(x|y)$ are defined. In the example below, these probabilities are gaussian and can be sampled with standard methods. 

The quality of the approximation can be monitored by the degree to which the obtained kernel $K'$ is not symmetrical. The final value of the kernel is obtained as the average $K_{ij} = \frac 1 2 (K_{ij}' + K_{ji}')$.

\section{Tracelessness constraint}

The linear map $\mc N_\rho^* \mc N$ always has one trivial eigenvector of eigenvalue $1$ corresponding to a change in state normalization, because
\begin{equ}
\mc N_\rho^* \mc N(\rho) = \rho.
\end{equ}
But, in principle, $\rho$ is not within our tangent space, as it is not traceless.
In fact, $\rho$ is the component orthogonal (in the information metric) to the space of traceless operators. Hence we can automatically restrict our analysis to the valid traceless operators by subtracting the orthogonal projector to the component $\rho$ from $\mc N_\rho^* \mc N$.

At the level of the components $K_{ij}$, this yields the new matrix
\begin{equ}
\begin{split}
\tilde K_{ij} &= K_{ij} - \frac 1 n \ave{\delta_{x_i},\rho}_\rho \ave{\rho,\delta_{x_j}}_\rho
= K_{ij} - \frac 1 n.
\end{split}
\end{equ}
This has the same eigenvectors, except that the eigenvector $\rho$ (which has components all equal to $1/\sqrt{n}$) has eigenvalue zero instead of $1$.

\section{Principal component analysis}

The above algorithm is very similar to kernel PCA, with the difference that the kernel depends on the unlabeled data: it is learned. In kernel PCA, one would use in a similar manner the matrix $K_{ij} = f(x_i,x_j)$ for various functions $f$. For instance, one may use the gaussian function 
\begin{equ}
\label{gaussian}
f(x,y) = e^{-\frac 1 {2 \delta^2}\|x-y\|^2}
\end{equ}
(RBF kernel) with the parameter $\delta$ optimized over some objective function. 
A problem is that there are generally no rules on how to choose $f$.

\section{Example}

\begin{figure}
\begin{tabular}{c}
\includegraphics[width=1\columnwidth]{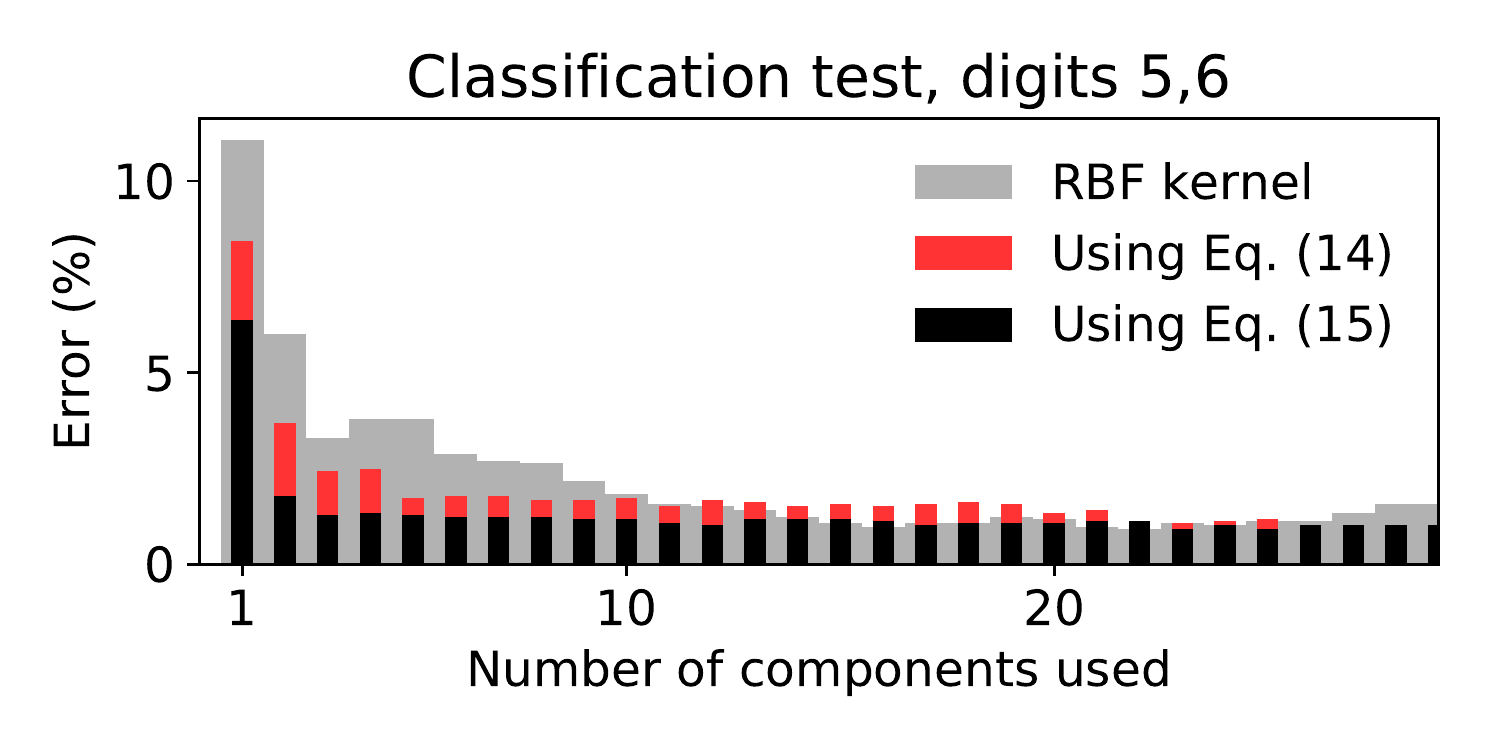}\\
\includegraphics[width=1\columnwidth]{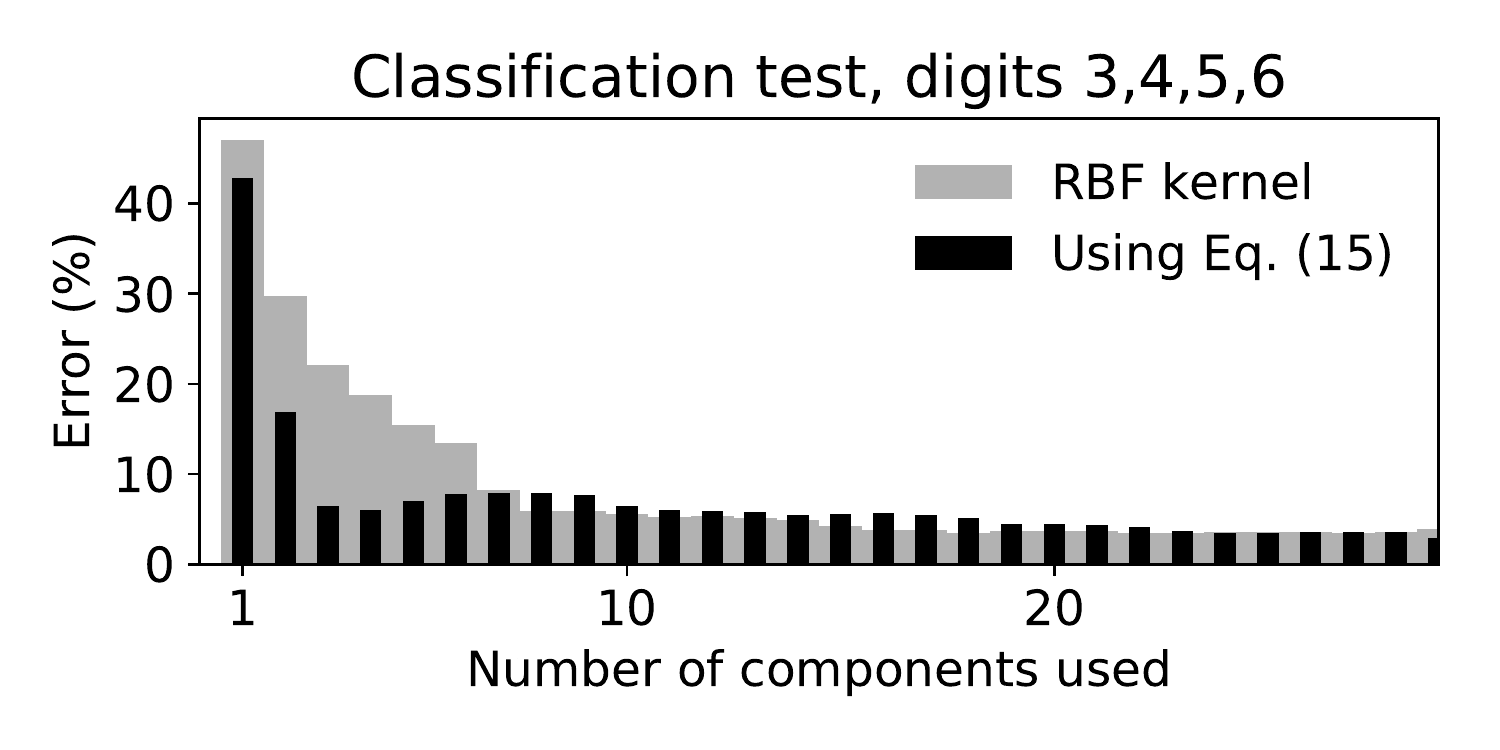}\\
\end{tabular}
\caption{
Classification test error probability as a function of the number of principal components used. The results from our approach are compared to a standard RBF kernel.
In order to test whether the principal components that we derived encode pertinent information about the images of handwritten digits, we used the label data to classify test images using the first $k$ principal components (in decreasing order of eigenvalues). These graphs display the error percentage in the classification task as a function of $k$. Because of the computational cost of our algorithm, we restricted the data to only a few digits: 5 and 6 in the first graph and 3,4,5,6 in the second. 
The classification algorithm simply uses voting from the 3 nearest labeled neighbors with votes weighted by the Euclidean distance in the space of $k$ components. We see that the first few features derived from our approach already encode most of the relevant information about the labels. 
An average of $1024$ training examples per digits were used in both cases because of computational constraints. 
Our kernel (Eq.~\eqref{kernel}) was evaluated using 140,000 gaussian samples, leading to an average asymmetry of about $2\%$ (see ``Numerical approximation'' section). We use parameters $\sigma=1$ (pixel) and $h=1.317$.
The sole parameter $\delta$ for the RBF kernel was set to yield the least error for $k=2$ in the $5,6$ case ($\delta=40$), and for $k=3$ in the $3,4,5,6$ case ($\delta=33$).
}
\label{classification}
\end{figure}

As an example, we consider the MNIST database. These are $28\times 28$ pixels gray-scale images of hand-written digits, all centered and normalized. In order to keep the computation time reasonable, we considered only four different digits (3,4,5,6), and 1024 training examples only per digits, which sets $n=4096$.

The channel that we use removes spatial information up to a scale $\sigma$ (measured in pixels), and add local noise with a variance of $h$ (relative to a maximum grayscale value of $1$). If we write the images as functions $f,g$ mapping each pixel to its grayscale value, the transition probabilities are 
\begin{equation}
\label{transition}
p(f|g) \propto e^{-\frac 1 {h^2} \|f-A g\|^2},
\end{equation}
where the linear operator $A$ implements convolution with a gaussian of variance $\sigma$. 

In order to test the quality of the extracted features, we project all test images on the relevant features using either Eq.~\eqref{proj0} or Eq.~\eqref{proj}. The $k$ most relevant components of the labeled training examples and unlabeled test data are then used for classification of the unlabeled data, simply by using the label of the nearest-neighbors (using the Euclidean distance).

The result is plotted in Figure~\ref{classification}, where we used the parameters $\sigma = 1.0$ (pixels) and $h=1.317$. We see that the components obtained using Eq.~\eqref{proj} perform much better than Eq.~\eqref{proj0}, in addition of being much faster to evaluate.

The results are compared to features extracted with kernel PCA using the gaussian radial basis function (RBF) kernel. We see that, contrary to the results from the RBF kernel, the quality of the classification is already almost saturated with only a few components using our approach.

The choice of parameters $\sigma$ and $h$ can be explained as follows. From Eq.~\eqref{kernel}, it is clear that the training images $f_i$ enter the calculations only in their ``blurred'' form $Af_i$, where $A$ implements convolution by a gaussian of variance $\sigma$. Hence, $\sigma$ cannot be too large or it may wash out important features of the data. However, a nonzero value of $\sigma$ does improve the classification results. In fact, such slight blurring of the images was already found experimentally to be advantageous in standard kernel PCA. Hence, for a fairer comparison, the same convolution was applied to the training images used for the RBF kernel calculations.

We find that the value of $h$, on the other hand, needs to be surprisingly large. The value we chose corresponds to a variance in pixel grayscale value of 130\% (although it rapidly reduces to a smaller variance once averaged over neighboring pixels, leading to an overall variance in intensity of 12\%). 
This is likely explained by the fact that, in order for this approach to yield non-trivial results, the noise introduced by the channel has to be large enough so that the image of the training data points under the channel have sufficient overlap (as probability distributions) with neighboring points. Indeed, if that is not the case, then the matrix $K$ is factors into a direct sum of independent blocks. 

For found that, indeed, for smaller values of $h$, the most relevant components end up having zero support on many of the training images. Surprisingly, this does not necessarily lead to bad classification results, but we fear that this might be an artifact of the limited number of digits and training samples that we are using.

\section{Discussion}

We derived from first principles a technique able to learn relevant features from unlabeled data, which is a form of kernel PCA but with a kernel that depends on the training data. 
The approach we used was originally introduced to understand the information-theoretic aspects of renormalization in quantum field theory, hence it could provide a starting point for comparing renormalization and machine learning, and possibly merging techniques from the two fields. 

The resulting algorithm is magnitudes slower than kernel PCA in its current form. However, it is possible that it could be made useful through further optimization, or be used to prove the ``correctness'' of simpler kernels. Indeed, contrary to kernel PCA, its parameters (in the form of the channel $\mc N$) have a transparent interpretation, and the approach can be justified in information-theoretic terms.

A related advantage of our approach is that it is manifestly independent of the way the data is parameterized (or in fact even from the way the full probability distributions are parameterized). For instance, in standard PCA, one simply diagonalizes the empirical covariance matrix between parameters of the pure states. But under a change of coordinate, this matrix does not transform like that representing a linear map, and hence its eigenvalues are not invariant, nor its eigenvectors covariant. By contrast, $\mc N_\rho^* \mc N$ is a linear map, and its eigenvalues are meaningful. This problem is also present in Refs.~\cite{Machta2013,bradde2017}.

This is not to say that we completely understand why our approach seems to work in the context presented here.
What we do is clear: we are determining which modifications of the empirical state $\rho$ lose least distinguishably against the noise introduced by the channel $\mc N$. But why would the resulting features effectively characterize the various hidden classes within the dataset? In the example of handwritten digits, it could simply be the fact that the digits are specifically designed (through cultural evolution) so that they can be distinguished under visual noise of the kind characterized by our channel $\mc N$ (defined by the transition probabilities in Eq.~\eqref{transition}).

More broadly, this touches the question of precisely defining the task of unsupervised learning. Can it be defined as a general task given any dataset, or are more inputs needed to define it, such as some a priori decision about which aspects of the data may be important or not? 

In fact, all approaches to machine learning use some prior assumptions with little logical justification. 
For instance neural networks are an ansatz able to represent efficiently only a subset of possible probability distributions over the data. Those networks where designed using inspiration from the brain and trial and error. They probably function because their structure encode some unknown universal facts about the kind of data that they are used on.

Although the role that the channel $\mc N$ plays in renormalization is more straightforward (it represents experimental limitations such as a bound on the energy achievable), a question similar to the above also exists, as the exact nature of the coarse-graining used is rarely examined nor justified.

The present work could provide a firm starting point for exploring these questions. 

\vspace{0.5cm}

\mysection{Acknowledgments}

This work was supported by the ERC grants QFTCMPS and SIQS and by the cluster of excellence EXC 201 Quantum Engineering and Space-Time Research.

\bibliography{../complete.bib}

\end{document}